\documentclass[10pt,twocolumn,letterpaper]{article}

\usepackage{paralist}
\usepackage{cvpr}
\usepackage{times}
\usepackage{graphicx}
\usepackage{amsmath}
\usepackage{amssymb}
\usepackage{capt-of}
\usepackage{etoolbox}
\usepackage{adjustbox}
\usepackage{subcaption}

\usepackage{authblk}

\usepackage[breaklinks=true,bookmarks=false,colorlinks=true,pagebackref]{hyperref}

\cvprfinalcopy % *** Uncomment this line for the final submission

\def\cvprPaperID{****} % *** Enter the CVPR Paper ID here

\newcommand{\kaolin}{\emph{Kaolin}}

\begin{document}
%%%%%%%%% TITLE
\title{\vspace{-2cm} \emph{Kaolin}: A PyTorch Library for Accelerating 3D Deep Learning Research\\ \large 
{\normalfont \vspace{.3cm}  \textbf{\url{https://github.com/NVIDIAGameWorks/kaolin/}}} }

\makeatletter
\renewcommand\AB@affilsepx{, \protect\Affilfont}
\makeatother

\author[1,2]{ \vspace{-.5cm} Krishna Murthy J. \thanks{Equal contribution. Work done during an internship at NVIDIA.\\ Correspondence to krrish94@gmail.com, sfidler@nvidia.com}}
\author[$\ast$ 1,4]{Edward Smith }
\author[1]{Jean-Francois Lafleche}
\author[1]{Clement Fuji Tsang}
\author[1]{Artem Rozantsev}
\author[1,5,6]{Wenzheng Chen}
\author[1,6]{Tommy Xiang}
\author[1]{Rev Lebaredian}
\author[1,5,6]{Sanja Fidler }

\affil[1]{NVIDIA}
\affil[2]{Mila}
\affil[3]{Universit\'e de Montr\'eal}
\affil[4]{McGill University}
\affil[5]{Vector Institute}
\affil[6]{University of Toronto}

%%%%%%%%%%%%%%%%%%%%%%%%%%
% Splash figure
\makeatletter
\let\@oldmaketitle\@maketitle
\renewcommand{\@maketitle}{\@oldmaketitle
\vspace{-1cm}
\centering
\includegraphics[width=\textwidth]{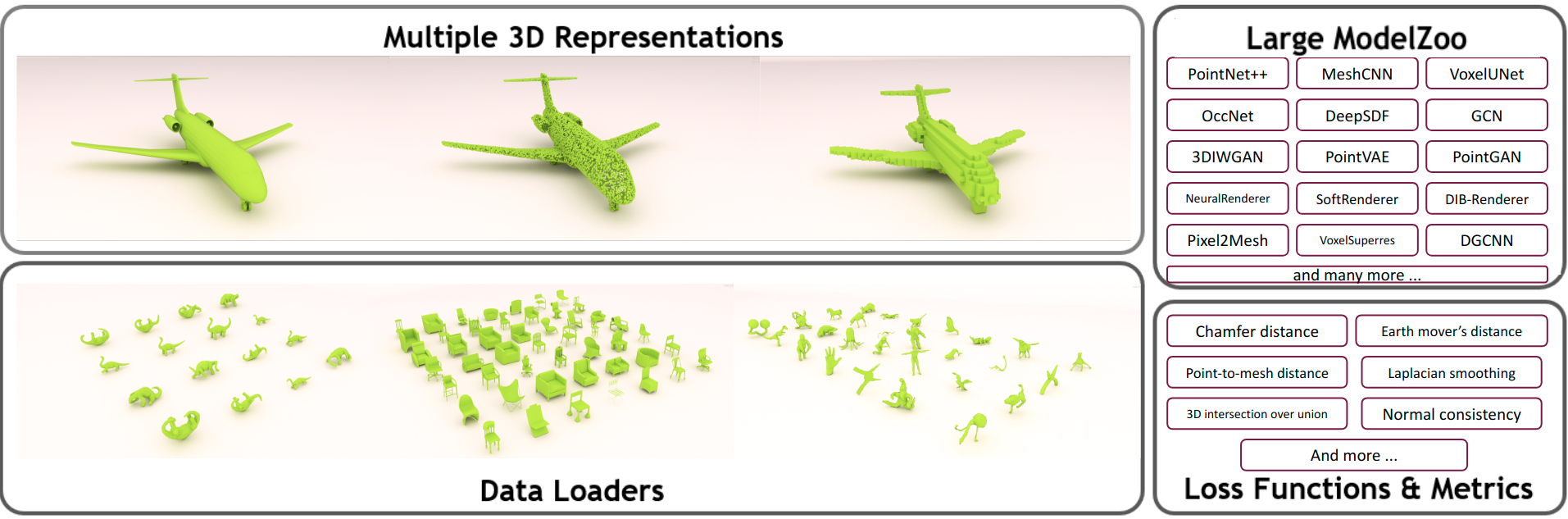}
\vspace{-5mm}
\captionof{figure}{\small 
{\sc Kaolin} is a PyTorch library aiming to accelerate 3D deep learning research. {\kaolin} provides {\bf 1)} functionality to load and preprocess popular 3D datasets, {\bf 2)} a large \emph{model zoo} of commonly used neural architectures and loss functions for 3D tasks on pointclouds, meshes, voxelgrids, signed distance functions, and RGB-D images, {\bf 3)} implements several existing differentiable renderers and supports several shaders in a modular way, {\bf 4)} features most of the common 3D metrics for easy evaluation of research results, {\bf 5)} functionality to visualize 3D results. Functions in {\kaolin} are highly optimized with significant speed-ups over existing 3D DL research code. 
% , making it an ideal starting point for researchers in this field.
}
\label{fig:splash}
\vspace{0.3cm}
}
\makeatother
% End splash figure
%%%%%%%%%%%%%%%%%%%%%%%%%%

\maketitle
%\thispagestyle{empty}

%%%%%%%%% ABSTRACT
\begin{abstract}
    We present \emph{Kaolin}\footnote{\kaolin{}, it's from Kaolinite, a form of plasticine (clay) that is sometimes used in 3D modeling.}, a PyTorch library aiming to accelerate 3D deep learning research. \kaolin{} provides efficient implementations of differentiable 3D modules for use in deep learning systems. With functionality to load and preprocess several popular 3D datasets, and native functions to manipulate meshes, pointclouds, signed distance functions, and voxel grids, \kaolin{} mitigates the need to write wasteful boilerplate code. \kaolin{} packages together several differentiable graphics modules including rendering, lighting, shading, and view warping. \kaolin{} also supports an array of loss functions and evaluation metrics for seamless evaluation and provides visualization functionality to render the 3D results.   Importantly, we curate a comprehensive \emph{model zoo} comprising many state-of-the-art 3D deep learning architectures, to serve as a starting point for future research endeavours. \kaolin{} is available as open-source software at \url{https://github.com/NVIDIAGameWorks/kaolin/}.
\end{abstract}

% Deep learning on 3D representations has seen significant interest over the last few years. While the interest in 3D learning is ever-increasing, there aren't any software packages geared towards accelerating research in this area. \emph{Kaolin} aims at doing presicely that! We provide efficient implementations of several differentiable layers that are commonly used in 3D deep learning applications. 

%------------------------------------------------------------------------

%%%%%%%%% BODY TEXT
\section{Introduction}

% \emph{There is this cool new paper on 3D on arXiv! You want to try it out. But they haven't released code! You still want to try it, but it will take you weeks to implement it -- you need to write data loaders, learn about a variety of 3D representations and graphics concepts, 3D-related loss functions, and evaluation metrics. And even when done, how do you visualize results? You feel it's better if you wait for a couple of months for the code release to come. If it comes.}

3D deep learning is receiving attention and recognition at an accelerated rate due to its high relevance in complex tasks such as robotics~\cite{Lalonde2006NaturalTC, yang2016real, sung2018robobarista, Yu2013AVR},  self-driving cars~\cite{Roddick2018OrthographicFT, mousavian20173d, chen2016monocular}, and augmented and virtual reality~\cite{garon2016real,2016ApplyingDL}. The advent of deep learning and an ever-growing compute infrastructures have allowed for the analysis of highly complicated, and previously intractable 3D data~\cite{3dconv, hamilton2017inductive, pointnet}. Furthermore, 3D vision research has started an interesting trend of exploiting well-known concepts from related areas such as robotics and computer graphics~\cite{NMR, MonteCarlo, milz2018visual}. Despite this accelerating interest, conducting research within the field involves a steep learning curve due to the lack of standardized tools. No system yet exists that would allow a researcher to easily load popular 3D datasets, convert 3D data across various representations and levels of complexity, plug into modern machine learning frameworks, and train and evaluate deep learning architectures. New researchers in the field of 3D deep learning must inevitably compile a collection of mismatched code snippets from various code bases to perform even basic tasks, which has resulted in an uncomfortable absence of comparisons across different state-of-the-art methods.

With the aim of removing the barriers to entry into 3D deep learning and expediting research, we present \kaolin{}, a 3D deep learning library for PyTorch~\cite{pytorch}. \kaolin{} provides efficient implementations of all core modules required to quickly build 3D deep learning applications. From loading and pre-processing data, to converting it across popular 3D representations (meshes, voxels, signed distance functions, pointclouds, etc.), to performing deep learning tasks on these representations, to computing task-specific metrics and visualizations of 3D data, \kaolin{} makes the entire life-cycle of a 3D deep learning applications intuitive and approachable. In addition, \kaolin{} implements a large set of popular methods for 3D tasks along with their pre-trained models in our \emph{model zoo}, to demonstrate the ease through which new methods can now be implemented, and to highlight it as a home for future 3D DL research. Finally, with the advent of differentiable renders for explicit modeling of geometric structure and other physical processes (lighting, shading, projection, etc.) in 3D deep learning applications~\cite{NMR, softras, dib}, \kaolin{} features a generic, modular differentiable renderer which easily extends to all popular differentiable rendering methods, and is also simple to build upon for future research and development.

% Many recent 3D deep learning applications make extensive use of differentiable renderers, to explicitly model geometric structure and other physical processes (lighting, shading, projection, etc.) to learn better representations. Due to the potential high impact that differentiable renderers may lead to in the 3D vision research landscape, we implement a generic, modular differentiable renderer that is easily accessible to all levels of research experience, and is also simple to build upon for further research and development. We also present concrete examples of other popular differential rendering methods methods, re-implemented using our modular renderer; demonstrating that most existing research in this realm, falls within our framework.

\begin{figure}[htb!]
    \centering
    \includegraphics[width=\linewidth]{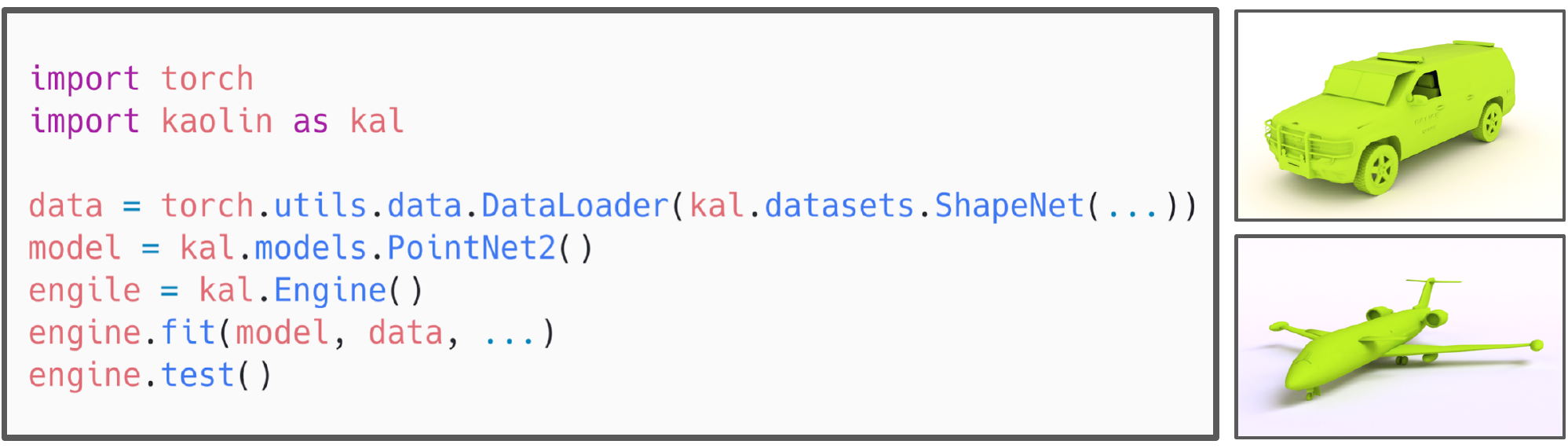}
    \vspace{-5mm}
    \caption{\kaolin{} makes training 3D DL models simple. We provide an illustration of the code required to train and test a PointNet++ classifier for \emph{car vs airplane} in 5 lines of code.}
    \label{fig:5loc}
\end{figure}

% Kaoling provides the following:
% \begin{compactitem}
%     \item Loading and pre-processing functionality for a number of 3D deep learning datasets including ShapeNet, PartNet, ScanNet, and many more. %, including a custom dataset released exclusively through the package. 
%     \item Full support for all popular 3D representations: meshes, point clouds, voxels, signed distance functions (level sets), and depth maps (RGB-D data). This includes easy conversion between all our representations.
%     \item Extensive functionality for 3D geometric and graphic operations. 
%     \item Efficient implementations of metrics for evaluating and training 3D tasks, across all representations.
%     \item Visualization functionality for all representations, including rendering options for headless applications. 
%     \item A fully modular differentiable rendering class, which is simple to build upon and easily extends to existing popular differentiable rendering methods. 
%     \item A large ModelZoo for popular 3D deep learning methods, across all our supported representations (Eg. PointNet~\cite{pointnet}, MeshCNN~\cite{meshcnn}, Pixel2Mesh~\cite{Pixel2Mesh}, Occupancy Networks~\cite{Occ}, and more).
%     %This includes 3 popular differentiable renderers, available to use out of the box. 
% \end{compactitem}

\begin{table*}[hbt]
    \centering
    \begin{adjustbox}{max width=\linewidth}
    \begin{tabular}{c|c|c|c|c| c}
         \emph{Library} & \# 3D representations & Common dataset preprocessing & Differentiable rendering & Model Zoo & USD support\\
         \hline
         GVNN~\cite{gvnn} & Mainly RGB(D) images & - & - & - & - \\
         Kornia~\cite{kornia} & Mainly RGB(D) images & - & - & - & - \\
         TensorFlow Graphics~\cite{TensorflowGraphicsIO2019} & Mainly meshes & - & \checkmark & \checkmark & - \\
         \hline
         \textbf{Kaolin (ours)} & \emph{Comprehensive} & \checkmark & \checkmark & \checkmark & \checkmark
    \end{tabular}
    \end{adjustbox}
    \caption{\kaolin{} is the first \emph{comprehensive} 3D DL library. With extensive support for various representations, datasets, and models, it complements existing 3D libraries such as TensorFlow Graphics~\cite{TensorflowGraphicsIO2019}, Kornia~\cite{kornia}, and GVNN~\cite{gvnn}.}
    \label{tab:compare}
\end{table*}

\begin{figure*}[!t]
    \centering
    % \begin{subfigure}[t]{0.3\textwidth}
    %     \centering
    %     \includegraphics[width=\textwidth]{figures/airplane_mesh_sky.png}
    %     \caption{Meshes}
    % \end{subfigure}%
    % ~ 
    % \begin{subfigure}[t]{0.3\textwidth}
    %     \centering
    %     \includegraphics[width=\textwidth]{figures/airplane_points_sky.png}
    %     \caption{Pointclouds}
    % \end{subfigure}
    % ~ 
    % \begin{subfigure}[t]{0.3\textwidth}
    %     \centering
    %     \includegraphics[width=\textwidth]{figures/airplane_voxels_sky.png}
    %     \caption{Voxel grids}
    % \end{subfigure}
    \includegraphics[width=\textwidth]{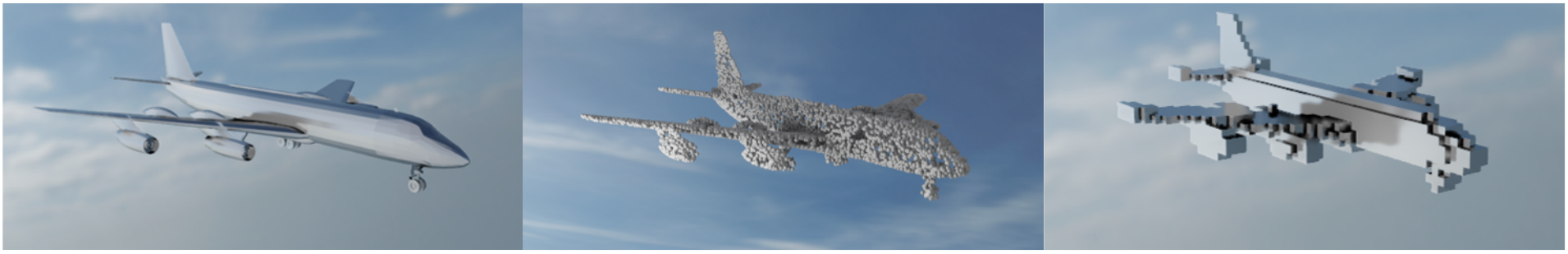}
    \caption{\kaolin{} provides efficient PyTorch operations for converting across 3D representations. While meshes, pointclouds, and voxel grids continue to be the most popular 3D representations, \kaolin{} has extensive support for signed distance functions (SDFs), orthographic depth maps (ODMs), and RGB-D images.}
\end{figure*}

\section{\kaolin{} - Overview}

\kaolin{} aims to provide efficient and easy-to-use tools for constructing 3D deep learning architectures and manipulating 3D data. By extensively providing useful boilerplate code, 3D deep learning researchers and practitioners can direct their efforts exclusively to developing the novel aspects of their applications. In the following section, we briefly describe each major functionality of this 3D deep learning package. For an illustrated overview see Fig.~\ref{fig:splash}.

\subsection{3D Representations}

% A highly important aspect of learning over 3D data is the representation of data that is chosen. Each representation posses its own collection of pros and cons, and as a result a particular choice of data type can have a large impact on the effect, feasibility, and ultimate success of a project. In addition 3D vision/robotics researchers must invariably work with multiple representations of 3D data, and perform conversions across them whenever needed. Hence, Kaolin provides full support to the 5 main classes of representation types:  

The choice of representation in a 3D deep learning project can have a large impact on its success due to the varied properties different 3D data types posses~\cite{3dDLtutorial}.
% Each representation comes with its own set of trade-offs (usually memory, fidelity, etc.).
% However, for new practitioners, this choice is often restricted by easy access to third-party packages 
% As it stands, this choice is often dictated by the functionality available for each representation or the original representation of the data, instead of their true trade-offs.
% To reduce these unnecessary restrictions, \kaolin{} provides full support for all popular 3D representations: 
To ensure high flexibility in this choice of representation, \kaolin{} exhaustively supports all popular 3D representations: 

\begin{compactitem}
    \item Polygon meshes
    \item Pointclouds
    \item Voxel grids
    \item Signed distance functions and level sets
    \item Depth images (2.5D)
\end{compactitem}
Each representation type is stored a as collection of PyTorch Tensors, within an independent class. This allows for operator overloading over common functions for data augmentation and modifications supported by the package. Efficient (and wherever possible, differentiable) conversions across representations are provided within each class. For example, we provide differentiable surface sampling mechanisms that enable conversion from polygon meshes to pointclouds, by application of the \emph{reparameterization trick}~\cite{GEOMetrics}. Network architectures are also supported for each representation, such as graph convolutional networks and MeshCNN for meshes\cite{gcn, meshcnn}, 3D convolutions for voxels\cite{3dconv}, and PointNet and PointNet++ for pointclouds\cite{pointnet, pointnet++}. The following piece of example code demonstrates the ease with which a mesh model can be loaded into \kaolin{}, differentiably converted into a point cloud, and then rendered in both representations:

{
\centering
\includegraphics[width=0.5\textwidth]{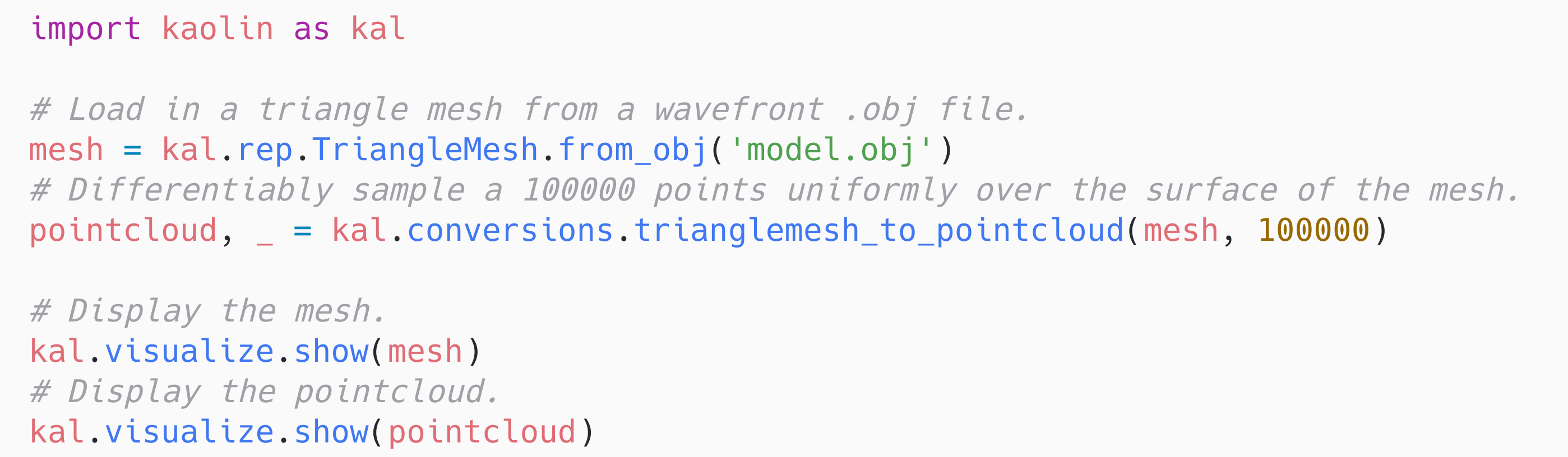}
}
% \caption{Converting across representations is as simple as a function call. Here, the \texttt{trianglemesh\_to\_pointcloud} function differentiably samples a pointcloud from the surface of a triangle mesh.}

% \begin{figure}[!h]
%     \centering
%     \includegraphics[width=0.5\textwidth]{figures/code_conversion.png}
%     \caption{Converting across representations is as simple as a function call. Here, the \texttt{trianglemesh\_to\_pointcloud} function differentiably samples a pointcloud from the surface of a triangle mesh.}
%     \label{fig:code_conversion}
% \end{figure}

\subsection{Datasets}
\kaolin{} provides complete support for many popular 3D datasets; reducing the large overhead involved in file handling, parsing, and augmentation into a single function call\footnote{For datasets which do not possess open access licenses, the data must be downloaded independently, and their location specified to \kaolin's dataloaders.}. Access to all data is provided via extensions to the PyTorch \texttt{Dataset}, and \texttt{DataLoader} classes. This makes pre-processing and loading 3D data as simple and intuitive as loading MNIST~\cite{MNIST}, and also directly grants users the efficient loading of batched data that PyTorch dataloaders natively support. All data is importable and exportable in Universal Scene Description (USD) format~\cite{usd}, which provides a common language for defining, packaging, assembling, and editing 3D data across graphics applications.

Datasets currently supported include ShapeNet~\cite{ShapeNet}, PartNet~\cite{partnet}, SHREC~\cite{ShapeNet, modelnet}, ModelNet~\cite{modelnet}, ScanNet~\cite{scannet}, HumanSeg~\cite{humanseg}, and many more common and custom collections. Through ShapeNet~\cite{ShapeNet}, for example, a huge repository of CAD models is provided, including over tens of thousands of objects, across dozens of classes. Through ScanNet~\cite{scannet}, more then 1500 RGD-B videos scans, including over 2.5 million unique depth maps are provided, with full annotations for camera pose, surface reconstructions, and semantic segmentations. Both these large collections of 3D information, and many more are easily accessed through single function calls. For example, access to ModelNet~\cite{modelnet} providing it to a Pytorch dataloader, and loading a batch of voxel models is as easy as: 

{
\centering
\includegraphics[width=0.5\textwidth]{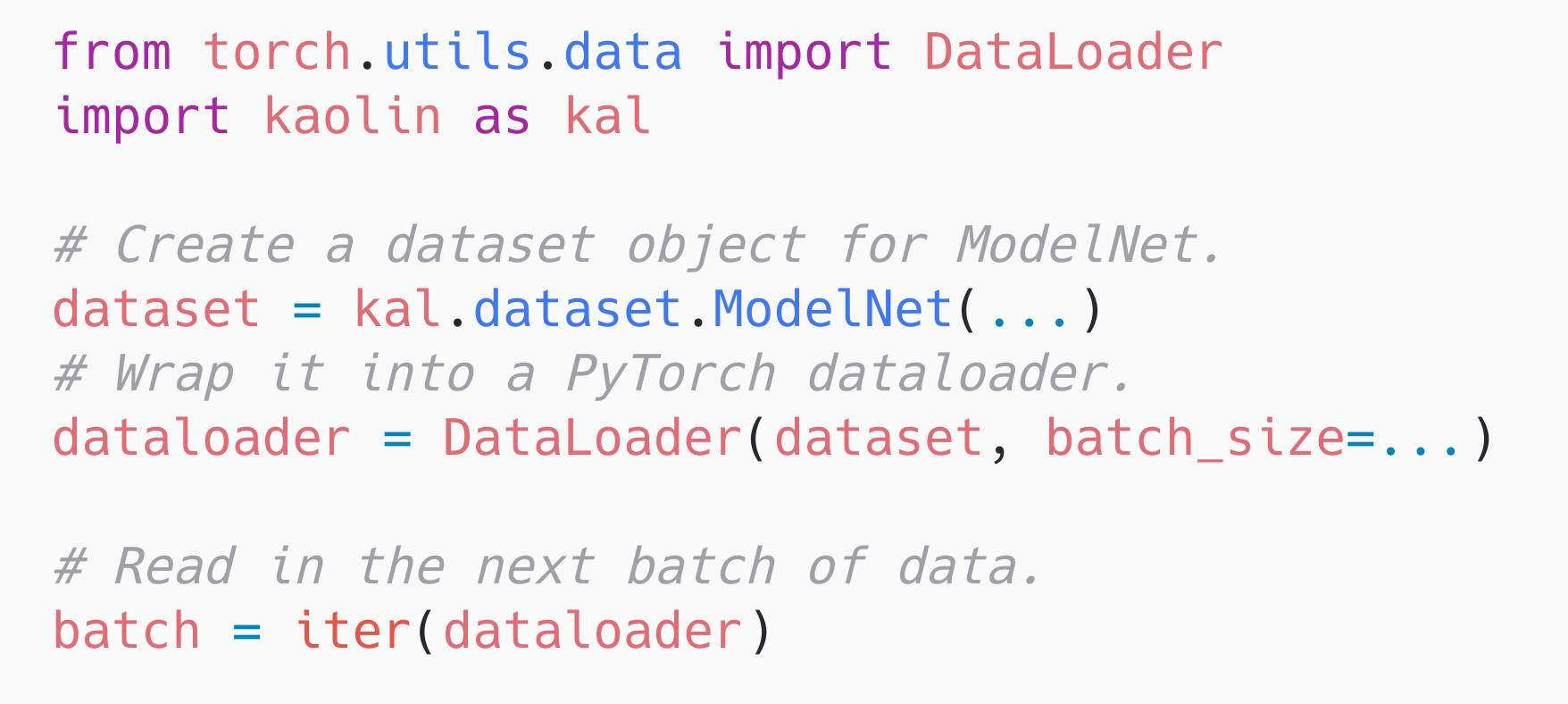}
}

% For datasets which do not possess open access licenses, the data must be downloaded independently, and their location specified to \kaolin's dataloaders.

\begin{figure}[!thb]
    \centering
    \includegraphics[width=0.5\textwidth]{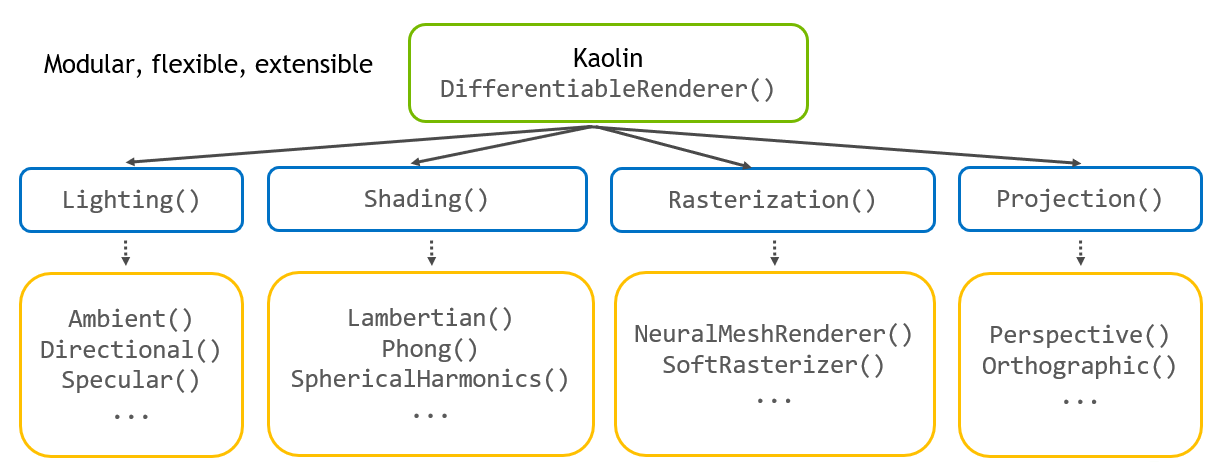}
    \caption{\textbf{Modular differentiable renderer}: \kaolin{} hosts a flexible, modular differentiable renderer that allows for easy swapping of individual sub-operation, to compose new variations.}
    \label{fig:dr_arch}
\end{figure}

\begin{figure*}[!th]
    \centering
    \includegraphics[width=\textwidth]{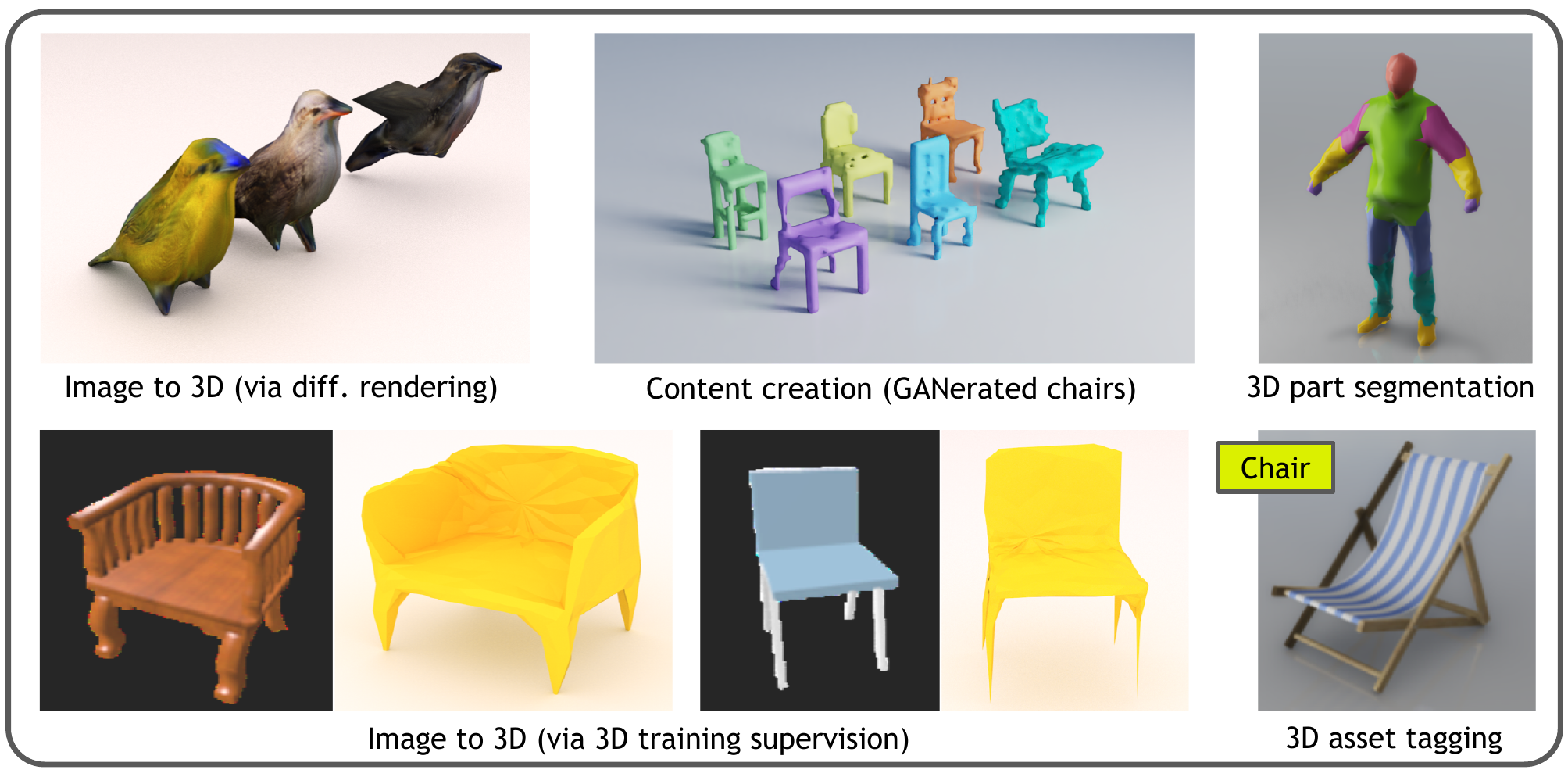}
    \caption{\textbf{Applications of \kaolin{}}: (Clockwise from top-left) 3D object prediction with 2D supervision~\cite{dib}, 3D content creation with generative adversarial networks~\cite{3DIWGAN}, 3D segmentation~\cite{meshcnn}, automatically tagging 3D assets from TurboSquid~\cite{turbosquid}, 3D object prediction with 3D supervision~\cite{GEOMetrics}, and a lot more...}
    \label{fig:applications}
\end{figure*}

\subsection{3D Geometry Functions}

At the core of \kaolin{} is an efficient suite of 3D geometric functions, which allow manipulation of 3D content. Rigid body transformations are implemented in several of their parameterizations (Euler angles, Lie groups, and Quaternions). Differentiable image warping layers, such as the perspective warping layers defined in GVNN (Neural network library for geometric vision) \cite{gvnn}, are also implemented.
The geometry submodule allows for 3D rigid-body, affine, and projective transformations, as well as 3D-2D projection, and 2D-3D backprojection. It currently supports orthographic and perspective (pinhole) projection.

\subsection{Modular Differentiable Renderer}

Recently, differentiable rendering has manifested into an active area of research, allowing deep learning researchers to perform 3D tasks using predominantly 2D supervision~\cite{NMR,softras,dib}. Developing differentiable rendering tools is no easy feat however; the operations involved are computationally heavy and complicated.
% due to the complicated nature and potentially huge computational burden of certain operation is must leverage. 
With the aim of removing these roadblocks to further research in this area, and to allow for easy use of popular differentiable rendering methods, \kaolin{} provides a flexible, and modular differentiable renderer. \kaolin{} defines an abstract base class---\texttt{DifferentiableRenderer}---containing abstract methods for each component in a rendering pipeline (geometric transformations, lighting, shading, rasterization, and projection). Assembling the components, swapping out modules, and developing new techniques using this abstract class is simple and intuitive.

\kaolin{} supports multiple lighting (ambient, directional, specular), shading (Lambertian, Phong, Cosine), projection (perspective, orthographic, distorted), and rasterization modes. An illustration of the architecture of the abstract \texttt{DifferentiableRenderer()} class is shown in Fig.~\ref{fig:dr_arch}. Wherever necessary, implementations are written in CUDA, for optimal performance (\cf{}~Table~\ref{table:timing}). To demonstrate the reduced overhead of development in this area, multiple publicly available differentiable renderers~\cite{NMR,softras,dib} are available as concrete instances of our \texttt{DifferentiableRenderer} class. One such example, DIB-Renderer~\cite{dib}, is instantiated and used to differentiably render a mesh to an image using Kaolin in the following few lines of code:

{
\centering
\includegraphics[width=0.5\textwidth]{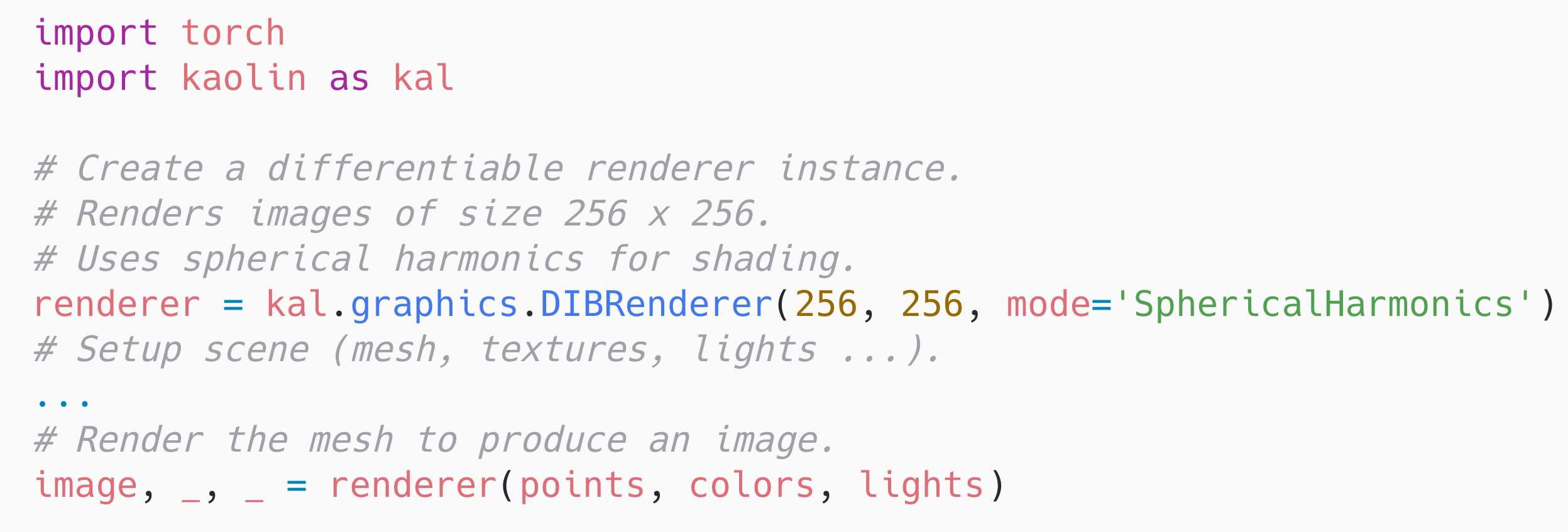}
}

% \TODO{Timing table pending, waiting on Clement's optimization.}

\subsection{Loss Functions and Metrics}

A common challenge for 3D deep learning applications lies in defining and implementing tools for evaluating performance and for supervising neural networks. For example, comparing surface representations such as meshes or point clouds might require matching positions of thousands of points or triangles, and CUDA functions are a necessity~\cite{fan2017point,Pixel2Mesh, GEOMetrics}. As a result, \kaolin{} provides implementations for an array of commonly used 3D metrics for each 3D representation. 
% Metrics are provided for each representation class, and wherever possible are implemented using CUDA for optimized performance. 
Included in this collection of metrics are intersection over union for voxels~\cite{3dr2n2}, Chamfer distance and (a quadratic approximation of) Earth-mover's distance for pointclouds~\cite{fan2017point}, and the point-to-surface loss~\cite{GEOMetrics} for Meshes, along with many other mesh metrics such as the laplacian, smoothness, and the edge length regularizers~\cite{Pixel2Mesh, NMR}.

\subsection{Model-zoo}

New researchers to the field of 3D Deep learning are faced with a storm of questions over the choice of 3D representations, model architectures, loss functions, etc. We ameliorate this by providing a rich collection of baselines, as well as state-of-the-art architectures for a variety of 3D tasks, including, but not limited to classification, segmentation, 3D reconstruction from images, super-resolution, and differentiable rendering. In addition to source code, we also release pre-trained models for these tasks on popular benchmarks, to serve as baselines for future research. We also hope that this will help encourage standardization in a field where evaluation methodology and criteria are still nascent.

Methods found in this model-zoo currently include Pixel2Mesh~\cite{Pixel2Mesh}, GEOMetrics~\cite{GEOMetrics}, and AtlasNet~\cite{atlasnet} for reconstructing mesh objects from single images, NM3DR~\cite{NMR}, Soft-Rasterizer~\cite{softras}, and Dib-Renderer~\cite{dib} for the same task with only 2D supervision, MeshCNN~\cite{meshcnn} is implemented for generic learning over meshes, PointNet~\cite{pointnet} and PointNet++~\cite{pointnet++} for generic learning over point clouds, 3D-GAN~\cite{3DGAN}, 3D-IWGAN~\cite{3DIWGAN}, and 3D-R2N2\cite{3dr2n2} for learning over distributions of voxels, and Occupancy Networks~\cite{Occ} and DeepSDF~\cite{deepsdf} for learning over level-set and SDFs, among many more. As examples of the these methods and the pre-trained models available to them in Figure \ref{fig:applications} we highlight an array of results directly accessible through Kaolin's model zoo.

\begin{table}[!b]
    \centering
    \begin{adjustbox}{max width=\linewidth}
    \begin{tabular}{c|c|c}
        \hline
        \emph{Feature/operation} & \emph{Reference approach} & \emph{Our speedup}  \\ \hline
        Mesh adjacency information & MeshCNN~\cite{meshcnn} & $110$X \\
        DIB-Renderer & DIB-R~\cite{dib} & $\sim 10$X \\
        Sign testing points with meshes & Occupancy Networks~\cite{Occ} & $> 10$X \\
        SoftRenderer & SoftRasterizer~\cite{softras} & $> 2$X\\ 
        \hline
    \end{tabular}
    \end{adjustbox}
    \caption{Sample speedups obtained by \kaolin{} over existing open-source code.}
    \label{table:timing}
\end{table}

% Figure \ref{fig:voxelgan_chairs} we demonstrate chair generation results produced using our 3D-IWGAN implementation, in Figure \ref{fig:human_seg} we demonstrate a segmentation of a human body produced using our MeshCNN implementation, and in Figure \ref{?} we demonstrate an objects reconstructed from a single images using our GEOMetrics implementation.

% \begin{figure}[!thb]
%     \centering
%     \includegraphics[width=0.5\textwidth]{figures/voxelgan_chairs.png}
%     \caption{A collection of chairs generated by 3D-IWGAN~\cite{3DIWGAN}.}
%     \label{fig:voxelgan_chairs}
% \end{figure}

% \begin{figure}[!thb]
%     \centering
%     \includegraphics[width=0.5\textwidth]{figures/human_seg.png}
%     \caption{MeshCNN \cite{meshcnn} segmenting a human body.}
%     \label{fig:human_seg}
% \end{figure}

\subsection{Visualization}

An undeniably important aspect of any computer vision task is visualizing data. For 3D data however, this is not at all trivial. While python packages exist for visualizing some datatypes, such as voxels and point clouds, no package supports visualization across all popular 3D representations. One of Kaolin's key features is visualization support for all of its representation types. This is implemented via lightweight visualization libraries such as Trimesh, and pptk for running time visualization. As all data is exportable to USD~\cite{usd}, 3D results can also easily be visualized in more intensive graphics applications with far higher fidelity (see Figure~\ref{fig:applications} for example renderings). For headless applications such as when running on a server that has no attached display, we provide compact utilities to render images and animations to disk, for visualization at a later point.

\section{Roadmap}

While we view \kaolin{} as a major step in accelerating 3D DL research, the efforts do not stop here. We intend to foster a strong open-source community around \kaolin{}, and welcome contributions from other 3D deep learning researchers and practitioners. In this section, we present a general roadmap of \kaolin{} as open-source software.

\begin{enumerate}
    \item \textbf{Model Zoo}: We seek to constantly keep improving our model zoo, especially given that \kaolin{} provides extensive functionality that reduces the time required to implement new methods (most approaches can be implemented in a day or two of work).
    \item \textbf{Differentiable rendering}: We plan on extending support to newer differentiable rendering tools, and include functionality for additional tasks such as domain randomization, material recovery, and the like.
    \item \textbf{LiDAR datasets}: We plan to include several large scale semantic and instance segmentation datasets. For example supporting S3DIS~\cite{S3DIS} and nuScenes~\cite{nuscenes} is a high-priority task for future releases.
    \item \textbf{3D object detection}: Currently, \kaolin{} does not have models for 3D object detection in its model zoo. This is a thrust area for future releases. 
    \item \textbf{Automatic Mixed Precision}: To make 3D neural network architectures more compact and fast, we are investigating the applicability of Automatic Mixed Precision (AMP) to commonly used 3D architectures (PointNet, MeshCNN, Voxel U-Net, etc.). Nvidia Apex supports most AMP modes for popular 2D deep learning architectures, and we would like to investigate extending this support to 3D. 
    % \Ed{Mention this is facilitated by Nvidia's deep learning SDK??}
    \item \textbf{Secondary light effects}: \kaolin{} currently only supports primary lighting effects for its differentiable rendering class, which limits the application's ability to reason about more complex scene information such as shadows. Future releases are planned to contain support for path-tracing and ray-tracing~\cite{MonteCarlo} such that these secondary effects are within the scope of the package. 
\end{enumerate}

We look forward to the 3D community trying out \kaolin{}, giving us feedback, and contributing to its development.

%------------------------------------------------------------------------

\section*{Acknowledgments}

The authors would like to thank Amlan Kar for suggesting the need for this library. We also thank Ankur Handa for his advice during the initial and final stages of the project. Many thanks to Johan Philion, Daiqing Li, Mark Brophy, Jun Gao, and Huan Ling who performed detailed internal reviews, and provided constructive comments. We also thank Gavriel State for all his help during the project.

{\small
\bibliographystyle{ieee_fullname}
\bibliography{egbib}
}

\end{document}